\documentclass{article}

\usepackage{PRIMEarxiv}

\usepackage[utf8]{inputenc} 
\usepackage[T1]{fontenc}    
\usepackage{hyperref}       
\usepackage{url}            
\usepackage{booktabs}       
\usepackage{amsfonts}       
\usepackage{nicefrac}       
\usepackage{microtype}      
\usepackage{lipsum}
\usepackage{fancyhdr}       
\usepackage{graphicx}       
\graphicspath{{media/}}     
\usepackage{tikz}
\usetikzlibrary{arrows}
\usepackage{ amsmath }
\newtheorem{Def}{Definition}
\DeclareMathAlphabet\mathzapf       {T1}{pzc} {mb} {it}
\usepackage{longtable}

\title{Diachronic Data Analysis Supports and Refines Conceptual Metaphor Theory
}

\author{
  Marie Teich\\
  Max Planck Institute for Mathematics in the Sciences\\
  Leipzig, Germany \\
  \texttt{marie.teich@mis.mpg.de} \\
   \And
  Wilmer Leal \\
  Max Planck Institute for Mathematics in the Sciences\\
  Bioinformatics Group, Department of Computer Science, Universität Leipzig \\
  Department of Computer Science, University of Florida, Gainesville, Florida \\
Gainesville, Florida \\
  \texttt{wilmer.leal@mis.mpg.de} \\
     \And
  J{\"u}rgen Jost\\
  Max Planck Institute for Mathematics in the Sciences\\
  Leipzig, Germany\\
      The Santa Fe Institute\\
      Santa Fe, New Mexico\\
  \texttt{jost@mis.mpg.de}
}

\begin{document}
\maketitle
\begin{abstract}
 As a contribution to metaphor analysis, we introduce    a statistical, data-based investigation with empirical analysis of long-standing conjectures and a first-ever empirical exploration of the systematic features of metaphors. Conversely, this also makes metaphor theory available as a  basis of meaning emergence that can be quantitatively explored and integrated into the framework of NLP.
\end{abstract}


\section{Introduction}

Metaphorically speaking, there is a wide river between computer-based data analysis methods on one side and Cognitive Linguistics (CL) and especially Conceptual Metaphor Theory (CMT) on the other side. Every now and then, a small boat attempts to cross this river, but our goal is to build a solid and lasting bridge. Less metaphorically, we need and want to address two different research communities simultaneously, each with its own concepts, ways of thinking and arguing, and discursive practices. To metaphor research, we present a statistical, data-based investigation that empirically analyzes long-standing conjectures and provides the first-ever exploration of the systematic structure underlying metaphors. To the Natural Language Processing community, we introduce metaphor theory as a  basis of meaning emergence that can be quantitatively explored and  whose understanding and integration into NLP methodologies hold great potential.

\subsection*{Cognitive Linguistics and Data Analysis}

Data driven linguistics is a very active and lively research field, but there exists a blind spot regarding the findings of Cognitive Linguistics \cite{tamari2020language,bisk2020experience} (CL for short).
CL is an actively developing branch of linguistics without an established closed canon. It depends on several widely accepted premises. The most important one is a unity of language and thought. This implies that language is not produced in a separate linguistic module of the brain,  but is instead processed and created by general cognitive mechanisms that are responsible for diverse cognitive tasks \cite{croft2004cognitive}. This has the important consequence that an understanding of linguistic structure offers a window into general principles of cognition.

Considering language in this way automatically links language research to many other areas of research. For example, with evolutionary anthropology that studies the development and role of human cognition and communication \cite{tomasello2005constructing}. More importantly for the focus of this work, however, it connects language to human embodiment, since meaning emerges in connection with and from the physical reality of humans, rather than in abstract isolation  \cite{rohrer2005image}.
Both of these points make computational analysis particularly challenging if its aim is to enable computers to "understand meaning" by analyzing the internal structures of language. The absence of the evolutionary functions of language and the absence of a body as well as of basal human experience mean that, from the perspective of Cognitive Linguistics, a computer can at best inadequately replicate the process of "understanding meaning". This is especially true for the most common computational linguistic method of natural language processing, where meaning of words is seen in its interrelations with other words according to the distribution hypothesis \cite{firth1957modes,harris1951methods,le2014distributed}. There, language is studied in isolation as an abstract system of signs, and text corpora are seen as containing meaning entirely within themselves.

Instead of trying to simulate the process of understanding, we want to keep our analysis within the framework of Conceptual Metaphor Theory. The mathematical structures we use do not have the purpose of imitating the structure of meaning, but only of enabling a corpus-based statistical analysis within the framework of an existing model.

\subsection*{Conceptual Metaphor Theory}

A central component of Cognitive Linguistics is Conceptual Metaphor Theory. This theory does not consider metaphor as a rhetorical device in the sense of garnishing meaning packaging. Rather, metaphor is the central cognitive linking mechanism between concrete embodied experiences and abstract linguistic domains.
Metaphor is not a simile, but a new transfer\textemdash hidden in its Greek root\textemdash  of experiential structures into abstract topics. Because of the linguistic nature of metaphor, abstraction can only be achieved in linguistic forms, and metaphor, as the instance of transfer, can be seen as the original point of new meaning-creation. 

Conceptual Metaphor Theory (CMT) emerged as a distinct field of research in linguistics with the publication of the book "Metaphors We Live By" by Lakoff and Johnson in 1980 \cite{lakoff2008metaphors}.
In that book the authors put forward several main theses and developed a terminology. Essentially, the claim is that metaphors are not only a phenomenon of poetic language, but are ubiquitous in everyday language.
Metaphoric transfers do not reside in individual rhetorical expressions, but systematically transfer structures between domains. With time, these mappings become part of our cognitive systems of thought and can give rise to a variety of individual metaphoric expressions. These systematic transfers are not comparisons of pre-existing structures, but constitute a model of the abstract domain in which it acquires its meaning.
Since its establishment, CMT has undergone a rich development and has found application in numerous fields such as politics, literature, pedagogy, and transcultural studies \cite{low2010researching}. However, a constant criticism concerns the lack of systematic research on the theory. The common research method of CMT studies particular language examples, supporting a separate thesis about the nature of metaphoric attributions.
The selection of these language examples is based entirely on the researcher's linguistic intuition along with their specific goal, which the example is intended to illustrate. This methodological issue is referred to as the first major problem of CMT \cite{kovecses2008conceptual}. To date, there is no empirical study that reconsiders the theoretical foundations of CMT based on real linguistic data.
This is particularly problematic because of the high flexibility of languages. A language forms a complex system with rules and tendencies on one side, but also expectations and competition between rules on the other side. Many phenomena can be discovered in such a system, but the appropriate tool to detect the guiding rules and main tendencies is statistical analysis.

There are attempts to perform corpus-based metaphor analysis with automated metaphor detection. Stefanowitsch and Gries \cite{stefanowitsch2007corpus} reviewed these attempts, but noted that in principle all attempts at automatic metaphor detection rely heavily on a presupposed metaphor theory. For example, one could search for typical metaphor source or target vocabulary or particular sentence structures, but this requires prior knowledge of the source and target distribution or of triggers indicating metaphors. Consequently, these methods are not immune to yielding highly biased results.

For this reason, this study takes a different approach to analyze a statistically significant set of randomly selected real-world metaphors. We chose to examine metaphors based on diachronic changes over the history of the English language.

\subsection*{Metaphor Research through Diachronic Changes}

Some frequently used metaphors experience an increasing conventionalization. They become so-called passive metaphors, where the original metaphoric use of the word is hardly recognizable as such \cite{bowdle2005career}. Typical examples of this stage are \textit{"Time is running"} or \textit{"Looking up to somebody"}. Continuation of this process can then lead to a new acquisition of word meaning or a transfer of meaning of a word from its source domain to its target domain. This type of exploration of metaphors has been successfully pursued by \cite{sweetser1990etymology}, for example.
The examination of metaphors through etymological mappings assumes a homogeneous process of meaning transfer and conventionalization across all relevant source and target domains, contingent solely upon the level of metaphorical language usage. If this is the case, no bias arises from the chosen method, and etymology, like archaeology for a historian, grants access to the study of active language.

\subsection*{Structure of the Paper}
The paper presents various methodologically distinct analyses of the data from the Mapping Metaphor project \cite{mappingmetaphor} exploring different facets of metaphor formation. By adopting the Mapping Metaphor project, we thereby necessarily accept and use the classification of metaphorical domains that is employed in that data set.
Metaphor formation is conceptualized as a network, where topics serve as nodes and edges represent metaphorical connections from sources to targets. First, we investigate the source and target topics of metaphors. Analyzing the distribution of connections among topics provides a global understanding of the dynamics of the metaphor network. A second set of experiments examines the network's division into two anti-communities of topics. This is further refined by a motif and a connection type number analysis. In the third part, we explore a characteristic feature of the network, namely, preferential attachment on edges. The fourth part focuses on the connection between metaphor and semantics by identifying roles within the metaphoric network. A final, fifth part, entails a discrete curvature analysis, which also leads to a critical role analysis of spatial metaphors.

\section*{Results}

\subsection*{Non-Randomness of Metaphorical Data} \label{non-randomness}

\begin{figure}
    \centering
    \includegraphics[width=.49\linewidth]{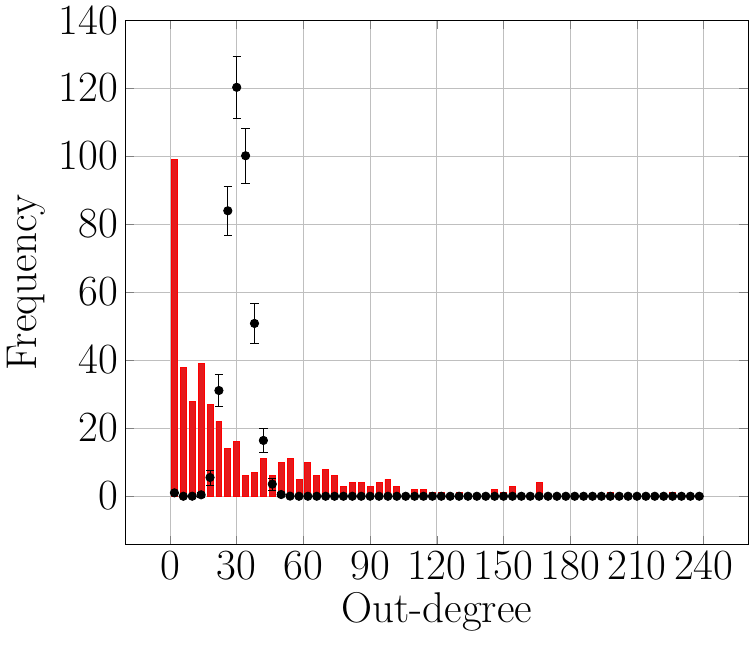}
    \includegraphics[width=.49\linewidth]{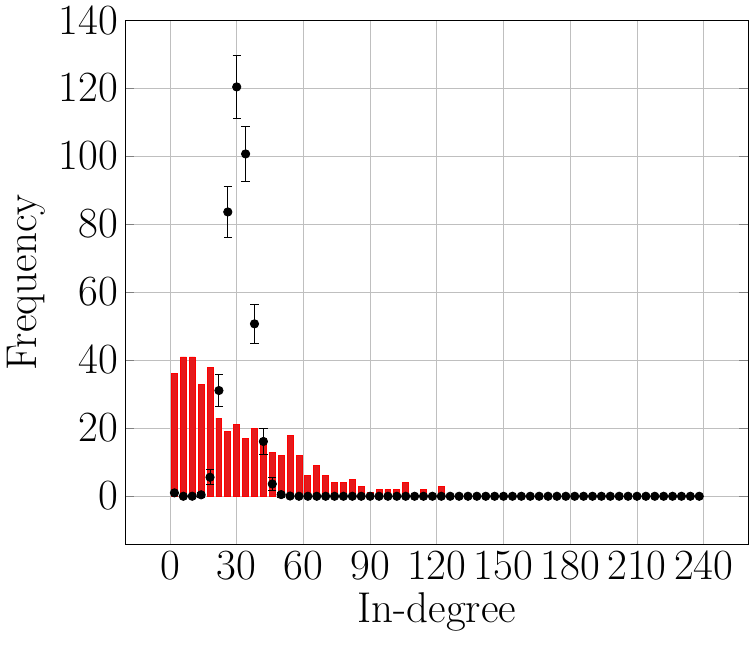}
    \includegraphics[width=.49\linewidth]{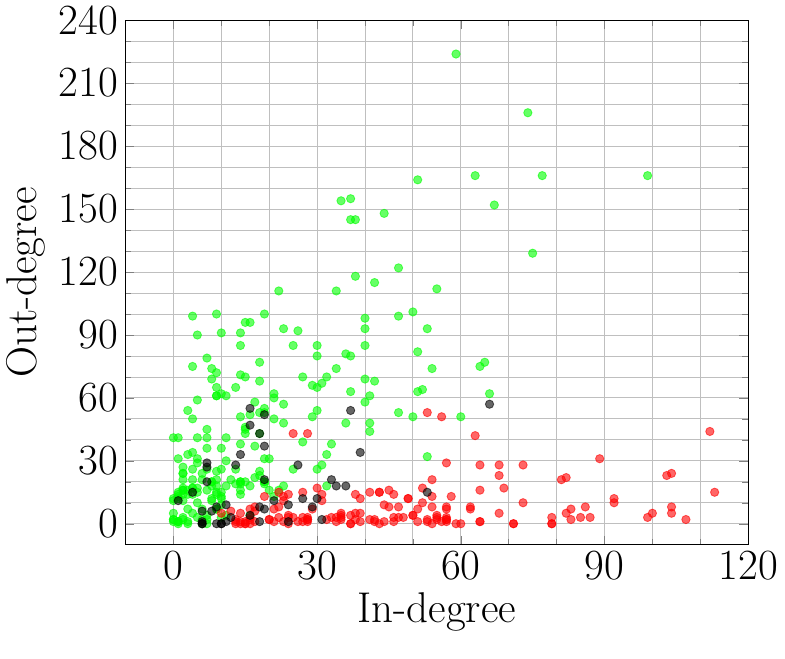}
    \includegraphics[width=.49\linewidth]{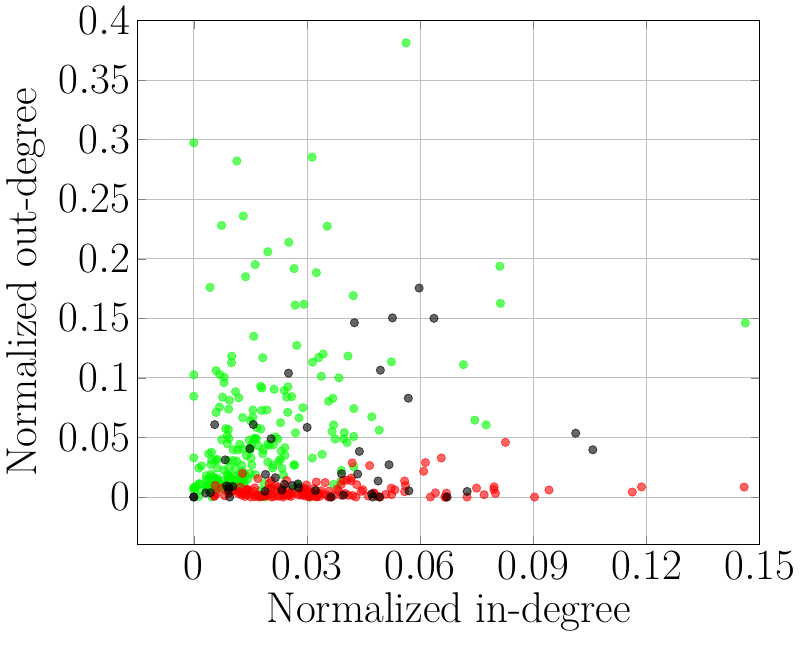}
    \caption{Top: Comparison of out-degree (left) and in-degree (right) distributions of categories in the metaphor network and configuration random model. Bottom: Absolute (left) and normalized (right) In- vs Outdegree of categories in the metaphor network}
    \label{fig:outdegree-dist}
\end{figure}

We first look at the in- and out-degree distribution of the directed graph representation of the Mapping Metaphor data (Figure \ref{fig:outdegree-dist}). The plots already include a comparison with the mean of one thousand equally sized Erdős-Rényi graphs (Materials and methods) and one standard deviation up and down. Right away, we find strong deviations of the in-degree and even stronger ones for the out-degree distribution of the metaphoric network from the random case. This already points to the interesting potential of the data for metaphor research, as non-randomness implies structure.

For the Erdős–Rényi model the degree distribution concentrates around the mean connection number of slightly above $30$ but for the metaphor data, this scale does not play a particular role. Instead, most categories cluster around low degree numbers and play only a minor role in the metaphor network. In contrast, there are a number of highly active categories, both for in- and out-degree, forming a so called heavy tail of the distribution. They show degree numbers several times higher than the degree maximally found in average random graphs.

\subsection*{Anti-communities of Concrete and Abstract Categories}

When we compare our metaphor graph with random graphs of the same size and average degree, we see striking differences and can identify important structural properties of the metaphor network. The latter's most marked structural property is that the majority of metaphors can be assigned to one of two classes that can be characterized by connectivity properties. The first class contains mainly spatial, mechanical, concrete and physically perceptible categories while the categories in the second class are mainly temporal, emotional and social. The two classes form anti-communities meaning that the majority of connections is between instead of inside the classes, although the first class also contains a significant number of internal connections. Figure \ref{fig:anticommunities_and_multiplicity_vs_number_connections}  shows that in comparison to configuration random graphs, there are two systematically dominating metaphor groups, one mapping from the first, more concrete anti-community to the second, more abstract one and another with mappings within the second anti-community, which shows the continuing re-determination and re-linking of concrete themes. The first group is the one most effectively described by CMT, while there is only little research and understand for the second group and it's role in conceptualization. 
The degree distribution (Figure \ref{fig:outdegree-dist}) is heavy tailed. In fact, most categories have low degrees and do not contribute much to the network, while some few  highly active categories have  in- or out-degrees that are significantly higher than the maximal degrees found in random networks. The most active sources and targets do not match, however, as seen in the plot. The vertices at the bottom right of the graph have a high target probability, but low or at most average source activity. In contrast, the strongest sources also have a high target activity. These two different activity groups -- one concentrated at the bottom right, the other on the diagonal --  in fact coincide with the two anti-communities identified above. Almost all categories on the diagonal, regardless of their degree,   belong to the concrete anti-community. The correlation coefficient for in- and out-degree of this set is $0.709$. In contrast, for the abstract anti-community it is only $0.273$, evidencing again the difference between the two classes.

\subsection*{Word Transfer Density}

In order to examine the word transfer density, we now consider the multigraph where each transferred word creates an edge between the corresponding categories. We then normalize the degrees by dividing by the total number of words in each category. Note, however, that this will in general underestimate the word transfer probability, in particular for very active mappings.\\
We use the same colors for the anti-communities as above (Figure \ref{fig:outdegree-dist}). With the expectation of the category "light" all topics with high in-degree density probability belong to the abstract anti-community and accumulate in the lower right corner, as before. The words in the other class, the concrete categories, the predominant sources, however, do not typically occupy a position on the diagonal, but are just active sources, but not preferred targets. So instead of a diagonal slope, we find that the probability of words in a category being involved in metaphorical mappings may be high either exclusively as a source or as target depending if the category belongs to the concrete or the abstract anti-community.

\subsection*{Motifs and the Metaphorical Mechanism} \label{motifs_result}

\begin{table}
    \centering
\begin{tabular}{|c|c|c|c|}
\hline
Metaphor mapping motif & z-value: overall & z-value: concrete
& z-value: abstract \\ \hline
\resizebox{0.18\textwidth}{!}{%
\begin{tikzpicture}[->,>=stealth',auto,node distance=3cm,
  thick,main node/.style={circle,draw,font=\sffamily\Large\bfseries}]
  \node[main node] (1) {A};
  \node[main node] (2) [right of=1] {B};
  \node[main node] (3) [right of=2] {C};
  \path[every node/.style={font=\sffamily\small}]
    (1) edge node [right] {} (2)
    (2) edge node [right] {} (3);
\end{tikzpicture}
}%
& 6.34
& 2.88
& 0.22 \\
\hline
\resizebox{0.18\textwidth}{!}{%
\begin{tikzpicture}[->,>=stealth',auto,node distance=3cm,
  thick,main node/.style={circle,draw,font=\sffamily\Large\bfseries}]
  \node[main node] (1) {A};
  \node[main node] (2) [right of=1] {B};
  \node[main node] (3) [right of=2] {C};
  \path[every node/.style={font=\sffamily\small}]
    (1) edge node [right] {} (2)
    (3) edge node [right] {} (2);
\end{tikzpicture}
}%
& 6.43 
& -0.55
& -0.73
\\ 
\hline
 \resizebox{0.18\textwidth}{!}{%
\begin{tikzpicture}[->,>=stealth',auto,node distance=3cm,
  thick,main node/.style={circle,draw,font=\sffamily\Large\bfseries}]
  \node[main node] (1) {A};
  \node[main node] (2) [right of=1] {B};
  \node[main node] (3) [right of=2] {C};
  \path[every node/.style={font=\sffamily\small}]
    (2) edge node [right] {} (1)
    (2) edge node [right] {} (3);
\end{tikzpicture}
}%
 & 6.44 
 & 0.42
 & -0.81
 \\
 \hline
 \resizebox{0.18\textwidth}{!}{%
\begin{tikzpicture}[->,>=stealth',auto,node distance=3cm,
  thick,main node/.style={circle,draw,font=\sffamily\Large\bfseries}]
  \node[main node] (1) {A};
  \node[main node] (2) [right of=1] {B};
  \node[main node] (3) [right of=2] {C};
  \path[every node/.style={font=\sffamily\small}]
    (1) edge node [right] {} (2)
    (2) edge node [right] {} (3)
    (3) edge[bend right] node [left] {} (1);
\end{tikzpicture}
}%
 & -2.46
 & -2.66
 & -0.65
 \\ 
\hline
 \resizebox{0.18\textwidth}{!}{%
\begin{tikzpicture}[->,>=stealth',auto,node distance=3cm,
  thick,main node/.style={circle,draw,font=\sffamily\Large\bfseries}]
  \node[main node] (1) {A};
  \node[main node] (2) [right of=1] {B};
  \node[main node] (3) [right of=2] {C};
  \path[every node/.style={font=\sffamily\small}]
    (1) edge node [right] {} (2)
    (2) edge node [right] {} (3)
    (1) edge[bend left] node [left] {} (3);
\end{tikzpicture}
}%
& -6.36
& -0.52
& 0.71
\\
\hline
\resizebox{0.12\textwidth}{!}{%
\begin{tikzpicture}[->,>=stealth',auto,node distance=3cm,
  thick,main node/.style={circle,draw,font=\sffamily\Large\bfseries}]
  \node[main node] (1) {A};
  \node[main node] (2) [right of=1] {B};
  \path[every node/.style={font=\sffamily\small}]
    (1) edge node [right] {} (2);
\end{tikzpicture}
}%
&  -9.53
& -7.21
& -2.92
\\
\hline
\resizebox{0.12\textwidth}{!}{%
\begin{tikzpicture}[<->, >=stealth',auto, node distance=3cm,
  thick,main node/.style={circle,draw,font=\sffamily\Large\bfseries}]
  \node[main node] (1) {A};
  \node[main node] (2) [right of=1] {B};
  \path[every node/.style={font=\sffamily\small}]
   (1) edge node [right] {} (2);
\end{tikzpicture}
}%
& 9.53 
& 7.21
& 2.92
\\
\hline
\end{tabular}
\caption{Non-symmetric three-Vertices motifs with corresponding z-value. Two-vertex motifs with corresponding z-value. For complete network, only connections within concrete topics block and only connections within abstract topic block.}\label{tab:network-motifs}
\end{table}

The motif count reveals further structure (Table \ref{tab:network-motifs}) . We find five subgraphs of size $n=3$ with Z-values either above $2$ or below $-2$, and in four cases, the absolute Z-value is even higher than $6$. The three motifs of size $3$ with one missing edge are clearly more frequent. In contrast,  closed triangles are almost absent in the data. Both effects, however, disappear when we restrict the analysis to the internal connections of the anti-communities. \\
  For motifs of size $2$, the deviation from randomness is even more pronounced, as symmetric connections dominate. This is true within both anti-communities, however to a much smaller extent within the abstract one. Accordingly, any metaphorical mapping from source to target makes the reverse mapping from target to source significantly more likely.

\subsection*{Persistence of Metaphorical Mappings} \label{persistence}

\begin{figure}
    \centering
    \includegraphics[width=.5\linewidth]{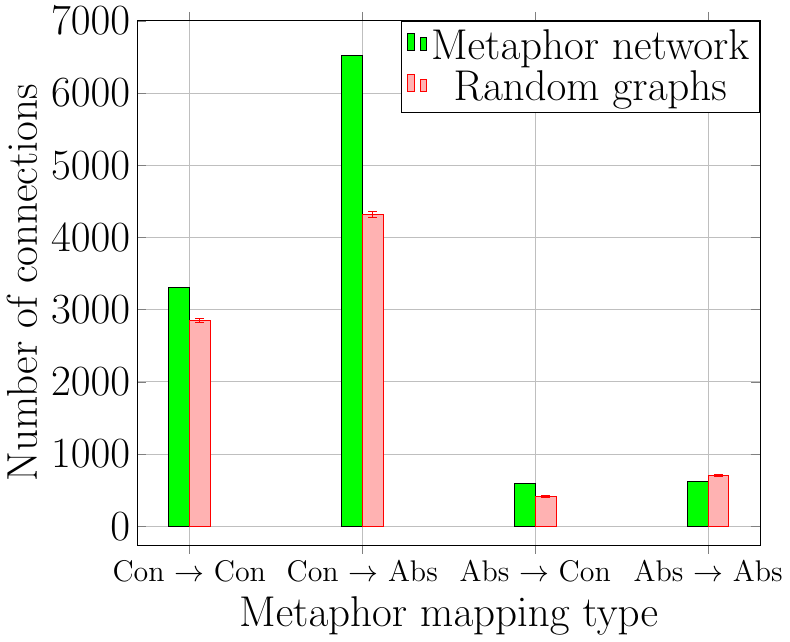}
    \includegraphics[width=.48\linewidth]{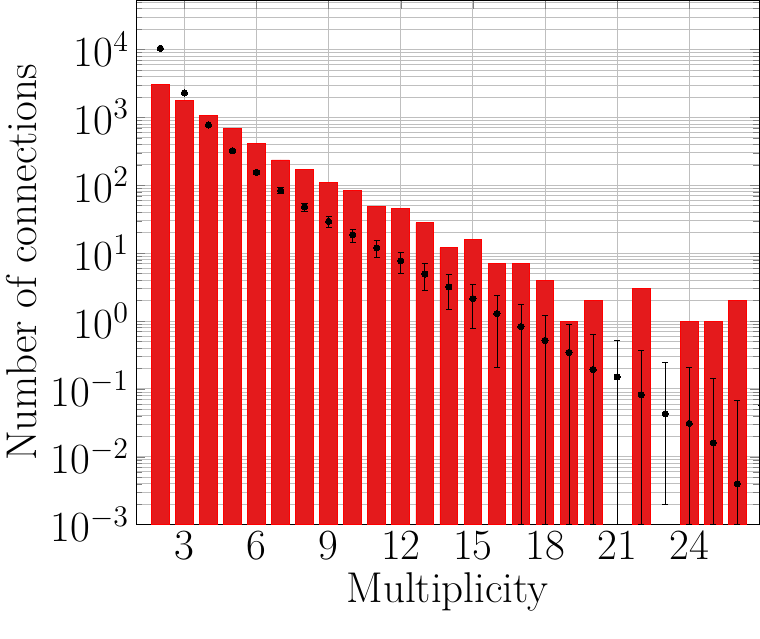}
    \caption{(Left) Connectivity patterns in the metaphoric mapping network distinguish two classes of categories that coincide with concrete categories (Con) and abstract categories (Abs). These classes form anti-communities, that is, most of the connections are between the classes and not within them. The bar chart compares the number of connections between classes observed in the metaphorical network (green bars) and the average number of connections observed in an ensemble of 1000 configuration random graphs (red bars with error bars). It shows that there are two systematically dominating metaphor groups, mappings from the concrete to the abstract anti-communities, and another with mappings within the concrete anti-community, which shows the continuing re-determination and re-linking of concrete themes. (Right) Comparison of multiplicity values of edges between configuration random model and the metaphor network. 
    }
    \label{fig:anticommunities_and_multiplicity_vs_number_connections}
\end{figure}

When looking at the multiplicity of edges, that is, the number of words transferred between categories, another deviation from a random model emerges (the black lines show the distribution for one thousand random configuration graphs with one standard deviation up and down for each data point.) As can be seen in Figure \ref{fig:anticommunities_and_multiplicity_vs_number_connections}, there is a significantly  higher probability of transferring words along already established mappings, so that vertex pairs with a high number of transferred words become significantly more frequent than in random models. \\
Such a deviation can always be a property of the methodology of the Mapping Metaphor data collection. It may also ultimately be the uneven distribution of a natural connection affinity between pairs of topics. The systematics of the deviation, however, primarily carries the properties of a preferential attachment mechanism on mappings, meaning that an already established metaphorical relation facilitates further word transfer along this relation.

\subsection*{Metaphor and Semantics} 

We now turn our attention to the role that categories play in the metaphor network, which is given by the connections they form.  In social network analysis, objects are considered to play the same role if they have the same incoming and out-going neighbors. To examine the roles within the metaphor network, we employed a Hierarchical Cluster Analysis (HCA) algorithm on the collection of categories and their incoming/outgoing neighbors. The distance between categories was measured by the number of non-common connections they possessed. The output of the HCA is a binary tree called dendrogram, whose clusters (branches) represent roles in the metaphor network. Real-world data sets often feature multiple equidistant objects, leading to numerous clustering solutions and consequently, a multitude of role schemes \cite{MacCuish2001,Leal2016}.  However, the HCA algorithm yielded only four dendrograms, one of which is illustrated in Figure \ref{fig:dendrogram}. Moreover, an overwhelming majority of roles (95.3\%) consistently emerge across all four dendrograms, underscoring the remarkably stable  clustering process and unveiling a robust underlying role structure within the metaphor network.

Through evaluating the metaphor distance as a similarity measure, we aim to assess its ability to provide insights into the semantics of categories. Comparing the metaphor-based similarity distance of all categories with cosine similarity and the euclidean distance derived from fasttext word embeddings reveals a low correlation. The cosine similarity yields a value of $-0.24$, while the euclidean distance results in $0.23$. Similarly, mutual information is present but not dominant, with values of $6.24$ for both cosine similarity and euclidean distance. However, the stability of the resulting dendrograms and the consistent ordering of categories vividly demonstrate that the metaphor-based role of a category encodes a semantic position, contributing to the formation of a semantic structure. This structure predominantly goes beyond the semantic information contained in word embeddings, encompassing figurative and conceptual semantic similarities. Consequently, while the acquired semantic structure does not completely diverge from word embeddings, it exhibits semantic aspects that surpass them, underscoring the limitations of word embeddings from a broader semantic perspective.

\begin{figure}
    \centering
    \includegraphics[width=\linewidth]{figures/categories_dendrograms_total_names_3.pdf}
    \caption{Hierarquical representation of semantic roles in the metaphor network: the figure is one of the four dendrograms obtained from running a Hierarchical Cluster Analysis algorithm on the set of categories from the metaphor network.  Categories were represented by their incoming $N_{\text{in}}$ and outgoing $N_{\text{out}}$ neighborhoods, the distance between categories $c$ and $c'$ was computed by counting the number of non-common connections: $d(c,c') = \frac{\#(N_{\text{in}}(c) \ominus N_{\text{in}}(c'))}{\#(N_{\text{in}}(c) \cup N_{\text{in}}(c'))} + \frac{\#(N_{\text{out}}(c) \ominus N_{\text{out}}(c'))}{\#(N_{\text{out}}(c) \cup N_{\text{out}}(c'))}$, and Ward's method was used as a grouping methodology.}
    \label{fig:dendrogram}
\end{figure}

\subsection*{The Role of Space and the Global Network Structure}\label{space}


A common and widely accepted assumption in the CMT literature is the specific role of spatial and directional domains for metaphor mappings. In the original set-up of the theory, Lakoff and Johnson elaborated on the so-called orientation metaphors. These do not structure or explain a single abstract concept, but rather perform the higher-level task of organizing the relations between different abstract concepts in terms of the spatial domain. Later, space was increasingly ascribed a special cognitive role in language. Spatial thinking was seen as the platform for schema formation - the universal building blocks of language. Thus, the spatial domain was considered to be the fundamental source of all concrete structures \cite{jackendoff1996architecture}. Such linguistic considerations went along with neuroscientific research about the role of spatial thinking in the hippocampus for memory and structural thought \cite{10.7554/eLife.16534} culminating in an actively pursued research program on cognitive spaces \cite{gardenfors2004conceptual,bellmund2018navigating}.

\begin{figure}
    \centering
    \includegraphics[width=.49\linewidth]{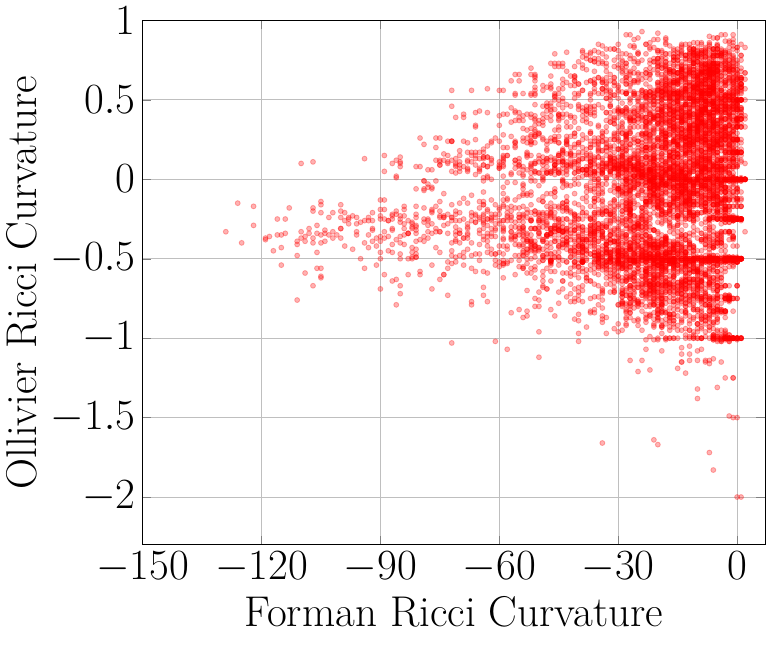}
    \includegraphics[width=.49\linewidth]{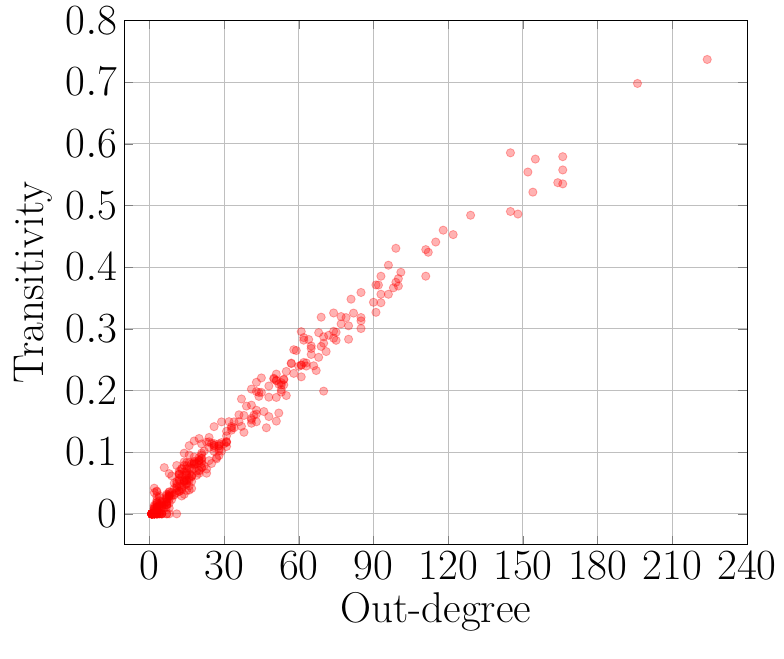}
    \caption{Distribution of Forman Ricci vs Ollivier Ricci curvature of edges in the metaphoric network (left).  Out-degree vs transitivity in the metaphor network (right)}
    \label{fig:Formand_vs_Ollivier_and_in_vs_outdegree}
\end{figure}

However, none of the analyses conducted on the mapping metaphor data showed a central, highlighted role of the spatial and directional categories in the metaphor network.
According to the theory of orientational metaphors, one might expect that mappings starting from spatial categories would influence their neighboring mappings, resulting in higher outward transitivity. In fact, the outwards transitivity of \textit{relative position} and \textit{movement in a specific direction}, is relatively high, but it does not exceed a general tendency found in the graph. This general tendency can be observed in the lower right figure \ref{fig:Formand_vs_Ollivier_and_in_vs_outdegree}, where one sees a clear correlation between the out-degree and the outwards transitivity of the vertices. Apart from being an interesting fact about the metaphorical data in itself, this correlation also shows that spatial regions follow this relation but do not exceed it. They do not play a special role in terms of their transitivity and thus their direct influence on their neighborhood in the network.

In the outlook of further possible special positions of the spatial domains and also with the aim of better understanding the general structure of the network, we consider two types of discrete curvature in the multigraph representation. The  statistics of local graph curvatures are known to be good indicators of global network properties. The Forman-Ricci curvature (see the appendix for details) captures the divergence or spreading in a network. Again, our analysis of the metaphor data show systematic differences from random graphs or other real world data sets. In particular, we see a rapid decrease at negative values, meaning that there are only very few edges with strong branching at their starting and/or terminal vertices (\ref{fig:Formand_vs_Ollivier_and_in_vs_outdegree}). Such edges would provide rapid flow access to many different metaphors, but this is definitely not the structure of the metaphor network.

\begin{figure}
    \centering
    \includegraphics[width=.49\linewidth]{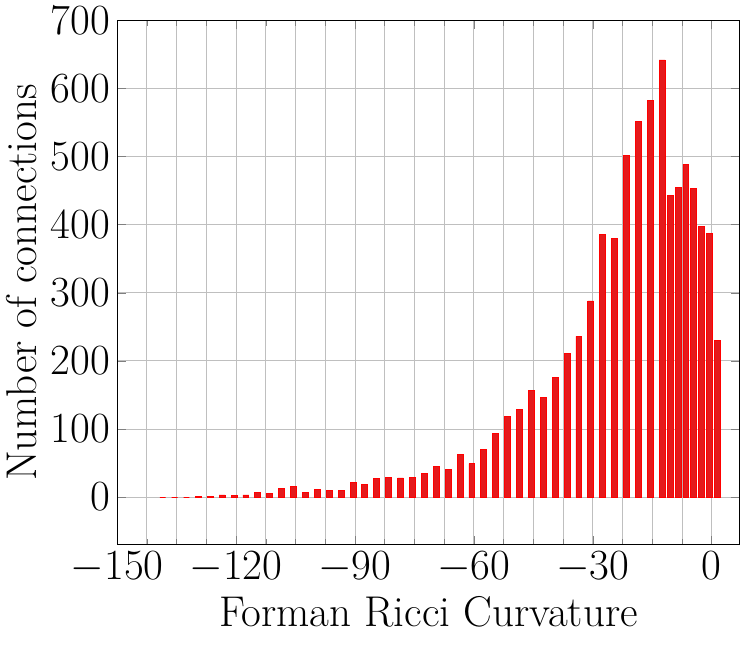}
    \includegraphics[width=.49\linewidth]{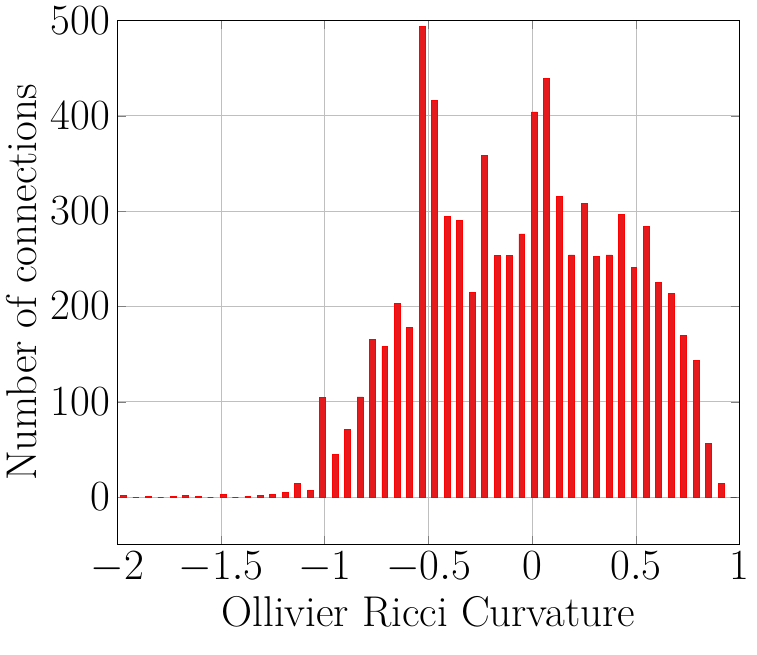}
    \caption{Distribution of Forman Ricci (left) and Ollivier Ricci (right) curvature of edges in the metaphoric network}
    \label{fig:Formand_and_Ollivier_Ricci_curvature}
\end{figure}

The Ollivier-Ricci curvature, in contrast, captures the presence of triangles and quadrangles in the network and thus the density of local neighborhoods. The existence of several seperated regimes of edges in the distribution indicates density clustering within the network on different scales. In the metaphor data, we do not find such dense local clusters separated from others as seen in (Figure  \ref{fig:Formand_and_Ollivier_Ricci_curvature}).

\section*{Discussion}

In this study, we have used a wide range of data analysis methods to systematically infer the properties of metaphor. We explore data of diachronic language change that resulted from metaphoric language use to draw back conclusions on the metaphoric language change itself. In doing so, we automatically rely on both the understanding of metaphors, which went into the data selection, and the categorization of the data.

On one hand, the data set that we have employed and the results derived from it support a basic assumption of CMT that metaphor is  a phenomenon of everyday language and that the choice of metaphor sources is influenced by experience and human embodiment. On the other hand, we also find global features of the metaphor network that have  not been observed before and sometimes even are at variance with some claims of CMT. Summarizing our findings, we find two classes that while forming anti-communities, behave quite differently with respect to their source and target activities. One class consists of the most active metaphor sources: simple and concrete topics such as 'Food and Eating', 'Shape' and 'Relative position'. Their source productivity is simultaneously correlated with a high target activity. The other, structurally very  different class contains the strongest metaphor targets. It contains mostly common but abstract topics such as 'Bad', 'Emotional Suffering' and 'Literature', which are little active as metaphors sources. 
 The word density of the sent and received words is  anti-correlated. This is very different from random null models. Active topics feature either a high density of sent or of received words. These effects are even stronger for the out-degree distribution than for that of the in-degrees.

Moreover, on the basis of the systematic data of the Metaphor Mapping project, we do not see any  special role of spatial and directional domains in the metaphor network, in contrast to a central claim of CMT. In fact, we do not see any category that would occupy a central position in the metaphor network, nor any hierarchies or clustering beyond bilateral edge relationships. Of course, it remains to be seen whether that can be confirmed by other data sets, but we are not aware of any other data set appropriate for a quantitative statistical analysis.

The division into the two anti-communities and the structure of the relationships between them seem to be the only clear global structure of the network. In agreement with an assertion of CMT,  mappings from the concrete anti-community to the abstract one play a prominent role. In addition, however, we also see many metaphorical relations between different concrete domains, indicating a rather fluid structure of that class. Even concrete terms do not make sense in isolation as their meaning conceptualization is embedded and actively renewed in a conceptual network. The consequences of this finding remain to be systematically explored and evaluated.

A review of the two anti-communities shows that the concrete one includes all bodily, spatial and geographic domains. The abstract topics include social constructions and spaces, emotion related domains and, remarkably, temporal domains. The latter is in agreement with the  philosophical conception of time as subjective time. Only the category 'period of time' belongs to the concrete anti-community.

And we have found that the metaphor network is globally strongly anti-transitive, as follows from the division into two anti-communities. Within each anti-community, however, such a strong anti-transitivity is not present, nor is the opposite, transitivity. From these findings, we conclude the mapping process is not triggered by a pre-existing structural analogy between two categories \cite{gentner1988metaphor}. The reason might be an omnipresent potential for analogy in general. But it rather seems in the opposite, that it is a tension between two categories that invites metaphors and thereby reveals structural similarities.

This is further confirmed by the high presence of symmetric metaphorical links between different domains. In more detail, there seem to be at least two possible explanations for the prevalence of symmetric connections. The restructuring of the target domain may make it more likely to become a source of a metaphor in the reverse direction. Or a metaphor may enable the target to  re-conceptualize the source domain. This feedback effect on the source may act like a seed for mappings in the opposite direction. That second possibility is supported by the global anti-transitivity that we have observed. The symmetrical component of metaphor surpasses the purely unidirectional understanding of metaphor in CMT.

The experiment in which the number of words per mapping is compared with configuration graph models (\ref{persistence}) provides, as far as we know, the first systematic and data-based evidence for persistent metaphoric mapping structures. It thereby confirms CMT's thesis that metaphoric connections are part of the cognitive structure, as opposed to the view that metaphors are individual rhetorical figures. To make this claim, we rely on the basic assumption of Cognitive Linguistics that the mechanisms of language are grounded in cognitive mechanisms. The experiment itself makes a statement on diachronic language evolution and shows that well-established mappings in a speech community also shape future language development. Within the paradigm of Cognitive Linguistics, we can conclude that the language we use, especially the well-established metaphorical mappings of that language, shapes our cognitive system and increases the likelihood of the emergence of new meanings along the typical mappings in our language.

The absence  of any flow or clustering or other emergent structure beyond bilateral connections suggests that there is neither a basic source domain from which most metaphors ultimately emerge nor a basic mechanism from the visual or any other domain that is the ultimate basis of conceptual thinking. Rather, there is quite a number of domains, related to different senses, that are productive as metaphor sources. We conclude from this the assumption that new metaphors can nevertheless emerge between every pair of topics with different probabilities following the tendencies of the network.

An important part of our research was devoted to the development of a semantic structure based on the metaphor network. Existing approaches to formalize a semantic structure are mainly word embeddings in the field of computational linguistics and, on the other hand, frame semantics in the framework of the Cognitive Linguistic approach \cite{fillmore2006frame}. Our approach is a substantial complement to these two existing approaches. Similar as in word embeddings, semantic distances are used, but as shown by the comparison experiments, the semantic distance defined by the role in the metaphor network is only weakly correlated and exhibits little mutual information with the semantic distances used in NLP. The crucial difference of our approach to frame semantics is that frame semantics does not provide a method to establish the relationship between topics organized in frames. We can fill precisely this gap by finding that the roles of each category created by all metaphor connections place the categories in a stable semantic net of relations. This net reveals not only the thematic, but above all the conceptual and figurative proximity between the categories.

There are many interesting examples of such figurative and conceptual proximity visible in the dendrograms presented. One of these is the proximity between \textit{dryness} with \textit{absence of life} on the one side and \textit{flowing and floating}, \textit{wetness}, and \textit{humidity} with \textit{life} on the other side. This double-sided proximity has no thematic share, but reproduces the deeply rooted conceptual proximity between fluidity and life, not only through direct metaphorical mappings, but even in an established parallel semantic structure.

Another interesting figurative aspect is revealed in the arrangement of emotion categories. There is a larger collection of strongly moving emotions: \textit{strong emotion and lack of emotion}, \textit{excitement}, \textit{pleasure}, \textit{anger}, \textit{emotional suffering}, \textit{pride}, \textit{contempt}. Additionally the categories \textit{intelligibility}, \textit{behaviour and conduct} and \textit{moral evil} found their way into this group of figuratively similar domains. But other emotions, that are figuratively less active, are not contained in this group, like \textit{fear}, \textit{esteem} and \textit{jealousy}. \\
Another characteristic observation of the results concerns different categories of movement or motion. These different categories are not grouped together, but on the contrary seem to mark a motif of different related groups. One motion motif is the category \textit{continuation} which finds itself in the group of categories: \textit{intention and planning}, \textit{increase in quantity} and \textit{command and control}, all controllable or constant processes. Similarly the category of \textit{slow action and degrees of caution} finds itself in a set of categories connected to thought processes and continuous change like \textit{completion} and \textit{cessation}. Another motion motif is \textit{repetition}, that is not unexpectedly grouped together with \textit{sky} and \textit{celestial sphere}. \textit{movement in a specific direction} is related to several spatial domains, like \textit{relative position}, \textit{size and spatial extent} and \textit{direction}. Finally \textit{progressive movement} finds itself close to \textit{impulse and impact} and is connected to the topics \textit{industry} but also of \textit{deity}, probably through conceptualizing the scheme of first causation and beginning.

This is only a glimpse of the discoveries that can be made by studying the acquired semantic structure from the perspective of different disciplines, such as psycholinguistics and psychology. Moreover, the metaphor role distance measure can also be used as a tool in discourse analysis. A change in framing can be classified as a more nuanced change if the old and new frames are from nearby categories, but is profound if the categories are not closely related in the semantic structure presented.

This leads to an important task for future research, to combine the information about semantic organization presented by the method of this work with the information provided by  word embeddings. This would enrich natural language processing limited through the distributional hypothesis by the figurative and conceptual aspect of semantic distances.

\section*{Conclusion}
We have provided a systematic and  multifaceted corpus-based metaphor analysis. Metaphors are novel acts of language usage that arise at a particular time, cause changes in conventional meaning and then can become conventionalized themselves. Therefore, data on diachronic language change can provide insight into this fundamental semantic mechanism.
The analysis empirically confirmed a basic assumption of conceptual metaphor theory, namely, that metaphors are persistent linguistic and cognitive transfer structures rather than discrete rhetorical figures.

The metaphor network encompasses a wide range of language topics, each exhibiting varying levels of activity. While the majority of topics demonstrate low activity, the activity is strongly concentrated on two groups of topics: on the one hand, concrete, sense-related domains, which are by far the richest sources but also serve as target domains for metaphors, and on the other abstract but nevertheless everyday topics, which are almost exclusively targets of metaphorical transmissions. This division is reflected in the global structure of the network. Most categories can be grouped into two anti-communities, a concrete and an abstract one. Along these anti-communities there are two systematically important metaphor processes. Mappings from the concrete to the abstract categories and mappings between concrete categories. Beyond that, no global structure is found, no clusters of thematic groups emerge that could be separated from each other. Also there is no flow of structure but only bilateral relations between two topics. Structurally, spatial metaphors do not occupy any special rank, they show a similar behavior as other concrete domains, which are not connected to the sense of orientation and vision.
The absence of any more particular structure in the network of metaphorical relations may suggest that this reflects a rather universal cognitive mechanism that can flexibly connect different domains and capture abstract structures through more concrete images, but at the same time also reorganizing the relations between concrete domains.

Our findings suggest that analogy or similarity is unlikely to be the primary selection criterion for new metaphor formation. Instead, they support the theory that metaphor is a creative process where the target domain is conceptualized through emerging similarity, exploiting the tension between the source and target domains. Additionally, there is evidence of feedback re-conceptualization of the source domain, indicating that metaphor is not a purely one-directional process.

Finally, it was found that the role of a topic in the metaphor network reveals in a stable way the semantic position of this topic relative to other topics. There is thus an explicit relation of the metaphor to semantic structure. We extracted the information on the semantic role from the metaphor network and constructed the resulting structure in form of a dendrogram. This structure does hold some mutual information as word embeddings based on the distributional hypothesis. At the same time, it contains information about conceptual and figurative semantic relations. This is a valuable and new addition to previously computed semantic structures, which allows for extensive further research.

\section*{Materials and Methods}\label{sec:materials-and-methods}

\subsection*{Mapping Metaphor Data}

We conducted our research based on the data of the Mapping Metaphor project \cite{mappingmetaphor,anderson2015mapping}, a collection of etymological transfers caused by metaphorical mappings.  The collection, conducted by the  School of Critical Studies of Glasgow University, is based on the Historical Thesaurus \cite{kay2009historical}, where each entry is assigned to one of the 415 thematic categories (e.g. \textit{movement}, \textit{visual arts}, \textit{weapons and armour}, etc.).

To detect metaphorical mappings, a computer program searches for words that appear in two or more categories. It is then manually inspected if these lexical overlaps can be seen as metaphorical usage. If this is the case, an entry is created with the transferred word, the source and the target category. The entry also includes the year in which the transferred meaning was detected for the first time as well as the last year of appearance if the usage has not been preserved until today. 
For our purposes it is important to note that the search for metaphorical mappings was systematically performed, and that not all existing etymological mappings have been included. In particular, since the project's main goal was to detect all possible connections between pairs of categories rather than each possible word transfer, less emphasis was placed on further findings along the same connection if several words had already been registered between a particular pair of categories.  Another relevant point is that the categorisation of the lexical items does not coincide with the notion of domain from Conceptual Metaphor Theory. Rather, most categories are broader and can be considered as associations of multiple, thematically close domains. Using the Mapping Metaphor data forces us to adopt its categorisation. Therefore we will mostly use categories in the sense of the Mapping Metaphor project, which allows only for a restricted transfer of the results into the CMT framework.

\subsection*{Mapping Metaphor Network}

As the data contains pairwise relations between source and target categories, it very naturally qualifies for a graph representation. A graph in fact represents a set of elements that can be pairwise related to each other. The crucial gain here is that graphs are mathematically very well studied objects for which a richly developed research toolbox already exists. Moreover, this toolbox can be adapted and extended to particular problems in the existing formalism \cite{newman2018networks}.

In the present paper we use two different graph representations of the data, as directed graphs or multigraphs, which will allow us to tackle different research questions. 

\subsubsection*{Directed Graph Representation}

When focusing not on single words, but on the network of mappings between the categories, we can use the following definition.

\begin{Def}
A directed graph $G=(V,E)$ consists of a set of vertices $V$ and a set $E$ of ordered pairs $(u,v)$ of vertices called directed edges. We will refer to them simply as edges. Furthermore, $u$ and $v$ are called the source and target of the edge $(u,v)$, respectively.
\end{Def}

In our application, the  vertices represent the  topic categories from the Mapping Metaphor data. If the data contains at least one word transferred from topic $u$ to topic $v$, this transfer gets represented by the edge $(u,v)$. This representation ignores the numbers of words transferred.  When we also want to include that information, we use the directed multigraph representation.
 
\subsubsection*{Directed Multigraph Representation}

Instead of focusing only on the categories as closed units and the connections between these units, we can consider  categories as forming sets of words.  Each transferred word then induces  an individual connection that should be represented by an edge. In this case our representation have to account for multiple edges having the same source and target.

\begin{Def}
A directed multigraph $G=(V,E)$ consists of a set of vertices $V$ and a multiset $E$ of ordered vertex pairs.
\end{Def}

Now, different words that are transferred from the same source to the same target are accounted for. Therefore, we do not loose the information about the number of words that were found.

\subsection*{In- and Out-Degree} \label{degree}


The degree is a basic concept of graph and network theory. It is defined for each vertex as the number of edges it participates in. For the directed graph and multigraph we distinguish between in- and out-degree. The out-degree, as the number of out-going connections, counts all edges in which a vertex appears as a source. Similarly, the number of edges which include a vertex as a target is the number of in-going connections to this vertex.

These simple notions allow us to identify local  characteristic properties of the individual categories. On the other hand, the statistics of their distributions also provide global information about the structure of the network, by comparing them with the corresponding distributions of random graph models. That is, we can infer to which extent our networks show, or not, specific structures and features present in random models. This will be a general principle in our graph analysis.

In the multigraph case, where the categories are not viewed as uniform entities,  but as word sets of different sizes, it is also interesting to look not only at the number of transferred words, which corresponds to the degree, but also at the transferred word density. This can be achieved by normalizing the in- and out-degree by the category size.

\subsection*{Random Graph Models}
\subsubsection*{Directed Erdős–Rényi Model} \label{er-model}

To get a first idea about the metaphoric network, that is the graph from the Mapping Metaphor data, we compare it with the most simple random directed graph model, the Erdős–Rényi model \cite{barabasi2013network}.

\begin{Def}
In the directed Erdős–Rényi model, a graph is chosen uniformly from the collection of all directed graphs with a given number of vertices $n$ and edges $N$. This means that all edges can appear with the same probability.
\end{Def}

\subsubsection*{Directed Configuration Model} \label{conf_model}

The configuration model describes random graph models with a prescribed degree distribution. In the directed case, not only the in- and out-degree distribution are fixed but also the degree sequences and by this way, the in- and out-degree combination for each vertex. The out-going and in-coming edge stubs then are randomly fused \cite{newman2018networks}.

\begin{Def}[Directed Configuration Model]
Let $n \geq 1$ and $N\geq 0$ be a fixed number of vertices and edges.  We denote by $i=(i_1, i_2, \ldots , i_n)$ and $o = (o_1, o_2,\ldots, o_n)$ the in- and out-degree sequences of the $n$ nodes.  It holds that $\sum^n_{v=1}i_v = \sum^n_{v=1}o_v = N$. To each vertex $v$, we attach $i_v$ in-going and $o_v$ out-going edge connections such that each in-going stub is matched with equal probability to each out-going stub. This way the probability for an edge to exist going from some vertex $v_1$ to $v_2$ is proportional to
\begin{equation}
    p_{v_1, v_2} \sim i_{v_1}\cdot o_{v_2}
\end{equation}
\end{Def}

The following algorithm implements a configuration model starting from a real world data model and keeping its degree sequences. \\
Two random connected pairs $ (v_1, v_2), (v_3, v_4)$ are cut and the connections are swapped, becoming $(v_1,v_4), (v_3,v_2) $. This step is repeated a given number of times $k$.  The output network approaches the directed configuration random multigraph.
One can also  prevent the swapping in case one of the edges $(v_1,v_4), (v_3,v_2) $ already exists. This modification of the algorithm makes the resulting network  a \textbf{directed configuration random graph} instead of a multigraph. The drawback is that one is no longer able to analytically compute the edge probability, but this modified algorithm will be important for some of our experiments presented to create comparable random graphs \cite{kashtan2005network}.

\subsection*{Anti-community Detection} \label{anti_community_detection}
The principle of anti-communities is exactly the opposite of clusters, to minimize the internal connectivity and maximize the connectivity across in- and outside of the group. Accordingly, the search can be performed using the negative graph in which disconnected nodes are connected to each other and vice versa. On this one of the well developed cluster detection algorithms can be applied. We use the Louvain cluster detection method based on greedy modularity maximization:
\begin{equation}
Q=\frac{1}{2 m} \sum_{i, j}\left[A_{i j}-\frac{k_i k_j}{2 m}\right] \delta\left(c_i, c_j\right),
\end{equation}
where $A_{ij}$ is the Adjacency matrix entry of the graph, $k_i$, the degree of node $i$, $m$ the total number of connections and $c_i$ the cluster of node $i$ while the Kronecker-delta function equals one only if $c_i = c_j$ and $0$ otherwise \cite{Blondel_2008, python_louvain}.
For each iteration, the algorithm produces slightly different results. In our case, the vast majority of categories are stably assigned to the same cluster. We run the assignment 10 times and keep only the overlapping sets as anti-communities while categories that change their assignment at least once are placed in the remainder set.

\subsection*{Edge Preferential Attachment} \label{edge_pref_attach}

In the directed multigraph case, we can not only count the in- and out-degree for each vertex, but we can also count the edges between each two connected categories in each direction. For our data, this represents the number of words that were found to be mapped from a specific source category to some specific target. 

The purpose of counting words along connections is to test whether metaphorical mappings do form persistent structures in language and mind, which is one of the main claims of Conceptual Metaphor Theory. In the Aristotelian view of metaphor as an individual rhetoric figure, individually mapped words should not influence the probability of other words being mapped along the same connection. In CMT, on the other hand, the phenomenon of metaphor is rooted not in the individual words and expressions, but in the conceptual transfer between two domains. Further, this transfer structure does not arise in the short term and temporally for production or processing of metaphoric expressions. Instead, it forms a persistent cognitive structure, which makes the transfer of many different partial aspects possible and becomes activated again and again. Once such a transfer structure gets established, the probability of further metaphoric expressions making use of this connection should increase.

The case of the random configuration model thus corresponds to the Aristotelian case. Each new formed edge of the multigraph is equally likely to form, given the degree sequences, and does not depend on previously existing identical edges. On the other hand, in the case of persistent mappings, the probability to form equal-directional edges between already connected vertices should increase.

It is possible to formulate exact models for which the probability of a new edge from $u$ to $v$ has a term proportional to the already existing number of edges along this possible connection:

\begin{equation}
    p_{u,v} \sim w_{u,v}
\end{equation}

and by comparison with the data, measure the influence of this term. However the collection method of the Mapping Metaphor data does not allow for this, as exploring further words along already detected connections has been systematically reduced. The aim of the project has not been to find all possible metaphorically transferred words but to focus on finding the mappings themselves. Thus, a quantitative examination of persistence is not possible. But we can test whether, despite the systematically lower word count, a confirmation of persistent metaphorical mappings can be found. This is done by comparing the distribution number of edges along a potential connection of the data against configuration random graph models (see \ref{persistence}).

\subsection*{Motif Analysis} \label{motif_analysis}

The analysis of graph motifs is a method for investigating the nature of processes behind the connections of a graph  \cite{alon2007network}. Graph motifs are defined in terms of small subgraphs that appear much more or much less often than expected from comparable configuration graphs. They reveal the logical properties according to which connections in a network emerge, and also those that do not play a role. We can then identify those different processes or mechanisms behind connection formation that  possess or lack these logical properties. \\
We will conduct a motif analysis on the network to test if the two such logical properties drive the formation of the metaphoric network, namely, transitivity and symmetry.

\subsubsection*{Transitivity} \label{transitivity}

Transitivity can be tested by looking at motifs of size $n=3$ that hold the following property:

\begin{equation}\label{eq:transitivity_prop}
     u\rightarrow v,\quad  v\rightarrow w \quad \Rightarrow \quad u\rightarrow w,
\end{equation}

where $u, v$ and $w$ are categories of the metaphoric network and $u\rightarrow v$ represents an edge between two categories $u$ and $v$. Transitivity is a logical property of every equivalence relation, for instance of analogy.

We computed the \textit{outward transitivity of a vertex} $u$ as the fraction of transitively closed directed triangles starting and ending at $u$, e.g., ($u\rightarrow v$, $v\rightarrow w$, $w\rightarrow u$). In other words, starting at one vertex, we consider the fraction of triangles that are transitively closed.

\begin{Def}[Outward Transitivity]
Let $G=(V,E)$ be a graph, $H_v$ the set of all vertices with an out-going edge from $v$, and let $K$ be the set of paths $(v\rightarrow h, h\rightarrow w)$ of length two, with $w\neq v$. The \textit{outward transitivity}  $T(v)$ of $v$ in $G$ is given by:

\begin{equation}
    T(v) = \frac{\#\{w\rightarrow v \in E : (v\rightarrow h, h\rightarrow w) \in K \}}{\#K}
\end{equation}
\end{Def}

\subsubsection*{Symmetry} \label{symmetry}

In a directed graph $G=(V,E)$ we can distinguish between symmetric and non-symmetric relations. A relation is non-symmetric if $u\rightarrow v$ is an edge of the graph, but $v\rightarrow u$ is not, while a relation is symmetric if both $u\rightarrow v$ and $v\rightarrow u$ are in $E$.  This property can be expressed as:

\begin{equation}\label{eq:symmetry}
    u\rightarrow v \quad \Rightarrow \quad v\rightarrow u.
\end{equation}

Consequently, symmetry can be quantified by computing the fraction of motifs of size $n=2$ that hold \ref{eq:symmetry}.  To find all motifs of size $n=2$ and $n=3$ in order to check the metaphorical mechanism for symmetry and transitivity, we make use of the motif detection software by \cite{kashtan2005network}.

\subsubsection*{Counting Motif Appearance}

Of course, when looking at our metaphorical mappings data, we cannot expect that transitivity and symmetry always hold, in particular since  the topic categories are rather broad and most probably combine several semantic domains so that different mappings can involve also quite distant aspects of one and the same category. Thus, we rather study tendencies, that is, a significant over- or under-representation of particular subgraph motifs when compared to random models. 

To find out which subgraphs are frequent or rare motifs of a network, we need the following definitions \cite{patra2020review}: 

\begin{Def}
A \textbf{subgraph} of $G=(V,E)$ is a graph $G'=(V',E')$ that consist of a subset $V'$ of $V$ and a connected subset $E'$ of edges  whose nodes belong to $V'$.  Furthermore, $G'$ is called an \textbf{induced subgraph} of $G$ if all edges $(u, v) \in E$ such that $u, v \in V'$, are also in $E'$.
\end{Def}

\begin{Def}
Two graphs $G=(V,E)$ and $G'=(V',E')$ are called \textbf{isomorphic} if there exists a bijection $f$ between $V$ and $V'$ that preserves adjacency, that is, the bijection $f$ is an isomorphism of $G$ and $G'$ if $(u,v) \in E$ implies that $(f(u),f(v))\in E'$.
\end{Def}

Isomorphic subgraphs represent of course the same motif. 
To count how many times a motif of $n$ vertices appear, one checks all non-identical connected combinations of $n$ vertices, including the overlapped ones. The exact counting algorithm can be viewed in \cite{kashtan2005network}. To assess the count $N_{\text {real }}\left(\Omega\right)$ and decide if a subgraph $\Omega$ is significantly over- or underrepresented in the metaphor network, it is compared with the count of the same subgraph in corresponding configuration graph models $N_{\text {rand }}\left(\Omega\right)$. The difference between the empirical network count and the mean random model count, $\left\langle N_{\text {rand }}\left(\Omega\right)\right\rangle$, is measured in units of standard deviation $\sigma\left(N_{\text {rand }}\left(\Omega\right)\right)$:

\begin{equation}
Z\left(\Omega\right)=\frac{N_{\text {real }}\left(\Omega\right)-\left\langle N_{\text {rand }}\left(\Omega\right)\right\rangle}{\sigma\left(N_{\text {rand }}\left(\Omega\right)\right)}.
\end{equation}
A subgraph is considered a significantly over- or underrepresented motif if $|Z\left(\Omega\right)| > 2$.

\subsection*{Similarity structure underlying the metaphoric network}
\label{sec:similarity-structure}
Two categories are structurally equivalent if they have the same incoming and out-going neighbors.  Sets of structural equivalent categories are said to form a position in the network, since they play the same role. In practice, though, sets of structurally equivalent vertices are rare and it is therefore necessary to relax this notion and to quantify how similar vertices are regarding their connections \cite{88504}. In order to identify roles within the metaphor network, we utilized Hierarchical Cluster Analysis (HCA), which is an unsupervised method. HCA takes a set of categories, along with their incoming and outgoing neighbors, as input, and generates a classification scheme. This scheme,  called dendrogram, is a binary tree that visually represents the underlying role structure of the categories in the metaphor network. HCA leverages the given metric structure of a dataset to construct the cluster hierarchy represented by the dendrogram. Agglomerative algorithms, in particular, follow an ascending grouping methodology. This process initiates with clusters consisting of single elements, referred to as leaves. At each subsequent step, the algorithm identifies the pair of closest clusters and merges them into a new cluster. The distances between this merged cluster and other clusters are then recalculated based on pairwise distances between the leaves, thereby progressively building the cluster hierarchy depicted in the dendrogram.

Clustering large data sets with, for example, distances limited to narrow ranges,  may produce several different dendrograms if two or more pairs of clusters are equidistant \cite{MacCuish2001}.  Therefore, it is necessary to consider these \textit{ties in proximity} and look for clusters, if any, that appear in a large fraction of the resulting dendrograms \cite{Leal2016}.

We run an HCA algorithm to classify categories by structural similarity in the network based on the method introduced in \cite{88504}. The set of categories is endowed with a metric structure by considering the following distance:

$$d(c,c') = \frac{\#(N_{\text{in}}(c) \ominus N_{\text{in}}(c'))}{\#(N_{\text{in}}(c) \cup N_{\text{in}}(c'))} + \frac{\#(N_{\text{out}}(c) \ominus N_{\text{out}}(c'))}{\#(N_{\text{out}}(c) \cup N_{\text{out}}(c'))}$$ \label{distance}
where $c$, $c'$ are categories, $N_{\text{in}}(c)$ and $N_{\text{out}}(c)$ are the incoming and outgoing neighbors of category $c$, and $N_{\text{in}}(c) \ominus N_{\text{in}}(c')$ is their symmetric difference.

Then, the Ward grouping methodology is used to produce all dendrograms resulting from ties in proximity. Cluster distances are updated  at each step according to the recursive Lance–Williams formula:

$$d(C_i\cup C_j,C_k) = \alpha_id(C_i,C_k) + \alpha_jd(C_j,C_k) + \beta d(C_i,C_j)$$
where $C_i, C_j$, and $C_k$ are disjoint clusters with size $n_{i},n_{j},$ and $ n_{k}$, respectively, and the coefficients are given by

$$\alpha_i = \frac{n_i + n_k}{n_i + n_j + n_k},\hspace{0.5cm} \alpha_j = \frac{n_j + n_k}{n_i + n_j + n_k}, \hspace{0.5cm} \beta = \frac{-n_k}{n_i + n_j + n_k}$$.

Then, we compute the fraction of dendrograms in which each of the resulting clusters appears, using the four contrast functions introduced in \cite{Leal2016}.

\subsection*{Fasttext Word Embedding} \label{word_embedding}

We want to investigate if the distances defined through the network neighborhood contain information similar or different to the semantic distance of a typical word embedding.  For this we look for correlations and mutual information between this network role distance and the embedding distances of words, using fastText word embedding in 300 dimensions. The word embedding is trained on Wikipedia and based on the skip-gram model described in \cite{bojanowski2017enriching}.

Typically, two different measures of semantic similarity and distance are used in word embeddings: the Euclidean distance between two words in space, or the cosine similarity between the two vectors from the origin to the words of interest. The Euclidean distance between points $x = (x_1, x_2, \ldots , x_{300})$ and $y = (y_1, y_2, \ldots , y_{300})$ in 300 dimensions is given by

\begin{equation}
    |x - y | = \sqrt{(x_1 - y_1)^2 + (x_2 - y_2)^2 + \ldots + (x_{300} - y_{300})^2}
\end{equation}

and the semantic dissimilarity according to the cosine similarity is computed as

\begin{equation}
    d_c = 1 - \cos(\vec{x}-\vec{y}) = 1 - \frac{\vec{x}\cdot\vec{y}}{|\vec{x}||\vec{y}|}.
\end{equation}

To test whether these two semantic measures are related to the metaphor network role similarity we not only look for possible correlations but also for possible mutual information \cite{scikit-learn}. The mutual information between two different measurements $U$ and $V$ is given by

\begin{equation}
M I(U, V)=\sum_{i=1}^{\#U} \sum_{j=1}^{\#V} \frac{\#(U_{i} \cap V_{j})}{N} \log \frac{\#(U_{i} \cap V_{j})\cdot N}{\#U_{i}\cdot \#V_{j}}
\end{equation}

where $\#U$ and $\#V$ are the number of samples. Compared to the linear correlation coefficient, the mutual information coefficient reveals in a much more general way the extent to which knowing the result of a measure on a data point informs the value of the second measure.

\subsection*{Graph Curvature} \label{curvature}

To understand the structure of the metaphor network in a more complete and general way, it is useful to learn about its "shape" just as we do to investigate geometric objects. In this section we design experiments to reveal the overall "shape" of the metaphor network.  The local geometry of an object is primarily described by its curvature and the statistics of those local observations informs about the global shape as a whole. This principle of local attributes that together allow inference on global properties was already applied when we looked at the degree distribution. But unlike the degree, discrete curvature is defined not for a vertex but for an edge \cite{eidi2020edge}, and therefore it directly informs about the structure of relations. There exist several  different translations of the continuous geometrical notion of curvature into discrete graph spaces. We shall look here into those two that have so far proved most useful in empirical data analysis, the Forman Ricci curvature and the Ollivier Ricci curvature \cite{samal2018comparative,Leal2020}. Both concepts have their origin in Riemannian  geometry, but allow us to study also structural properties in graphs that have no direct analogue in Riemannian geometry \cite{Leal2020}.

\subsubsection*{Forman Ricci Curvature} \label{forman}

The Forman Ricci curvature of a directed multigraph measures the prominence of flow structures in a network. For each connected pair of vertices $(v_i,v_j)$ it takes all uni-directed edges e.g. $e=(v_i,v_j)$ and counts the number of in going edges to its source $I_{v_i}$ and out going edges of its target $O_{v_j} $ thus quantifying to which degree this edges contributes to a flow structure. Following \cite{leal2021forman}, we define:

\begin{Def}[Forman Ricci Curvature]
Let $e$ be the non-empty set of edges from vertex $v_i$ to vertex $v_j$. Let further $\text{in}(v_i)$ be the set of in-going edges to $v_i$ and similarly $\text{out}(v_j)$ the set of out-going edges from $v_j$. Then the Forman Ricci Curvature of $e$ is defined as

\begin{equation}
    \mathbf{F}(e)= 2 - \# \text{in}(v_i) - \# \text{out}(v_j)
\end{equation}
\end{Def}

A more negative Forman Curvature indicates strong flow structure, while values around zero show the absence of flow.

\subsubsection*{Ollivier Ricci Curvature} \label{ollivier}
The Ollivier Ricci Curvature of an arc $(u,v)$ quantifies how close the incoming neighbors of $u$ are from the outgoing neighbors of $v$ \cite{ollivier2009ricci,samal2018comparative,eidi2020edge,Leal2020}.  Formally, 

$$\mathcal{O}(u,v) = \mu_0 - \mu_2 - 2\mu_3,$$

where $\mu_i$ quantifies the fraction of incoming neighbors of $u$ that are at distance $0\leq i \leq 3$ from $v$ in an optimal transport plan \cite{Eidi2020}.  Positively curved edges imply that a large fraction of incoming neighbors of $u$ are also outgoing neighbors of $v$, therefore they are involved in directed triangles $(x,u)(u,v)(v,x)$. Flat edges $(u,v)$ are, for instance, those whose incoming neighbors of $u$ are at distance one from the outgoing neighbors of $v$, forming directed quadrangles $(x,u)(u,v)(v,y)(y,x)$.  Negatively curved arcs are those for which the incoming neighbors of $u$ are at distance larger than one from the outgoing neighbors of $v$.

\section*{Acknowledgments}
The authors thank Eugenio Llanos from Corporación Scio for providing the Lisp packages to compute ties in proximity and cluster contrast to carry out Hierarchical Cluster Analysis.
We would also like to thank Wendy Anderson and the project team of Mapping Metaphor for providing the research data.

\bibliographystyle{unsrt}  
\bibliography{citations}  

\newpage

\section*{Appendices}

\noindent\textbf{Appendix A. Extended Material and Methods}\medskip

\noindent\textbf{Appendix B. Extended Results}\medskip

\appendix

\section{Extended Material and Methods}

\subsection*{Ollivier Ricci Curvature}

To define the Ollivier Ricci Curvature of an edge $e$ from $u$ to $v$, we compute an optimal transport plan between a probability measure $\mu_{\text{in}}$, defined on a neighborhood of $u$ called masses, and a probability measure $\mu_{\text{out}}$, defined on a neighborhood of $v$ called holes. The sets of masses and holes are defined as follows \cite{Eidi2020,eidi2020edge,leal2021forman}:

\begin{Def}[Masses and Holes of an Arc]
Given an arc $e=(u,v)$ of a directed graph $G=(V,E)$, we define the set of masses $\mathcal{M}$ of $e$ to be the singleton $\{u\}$ if $u$ does not have incoming neighbors. Otherwise, $\mathcal{M}$ is the set of incoming neighbors of $u$. Similarly, the set of holes $\mathcal{H}$ of $e$ is the singleton $\{v\}$ if $v$ does not have outgoing neighbors. Otherwise, $\mathcal{H}$ is the set of outgoing neighbors of $v$.
\end{Def}

The probability measures are defined as follows: If the tail $u$ of $e$ has no incoming neighbors, we set $\mu_{\text{in}}(u)=1$. Otherwise, for each mass $m \in \mathcal{M}$, we define $\mu_{\text{in}}(m)=\frac{1}{\#\text{in}(u)}$, where $\#\text{in}(u)$ is the number of incoming neighbors of $u$. Similarly, if the head $v$ of $e$ has no outgoing neighbors, we set $\mu_{\text{out}}(v)=1$. Otherwise, for each hole $h \in \mathcal{H}$, we define $\mu_{\text{out}}(h)=\frac{1}{\#\text{out}(v)}$, where $\#\text{out}(v)$ is the number of outgoing neighbors from $v$.

We can now define the Ollivier Ricci curvature of an arc $e=(u,v)$.

\begin{Def}[Ollivier Ricci Curvature]
Given an arc $e=(u,v)$ of a directed graph $G=(V,E)$, we define the Ollivier Ricci curvature $\mathcal{O}(u,v)$ as

\[
\mathcal{O}(e) = 1 - W_1(\mu_{\text{in}}, \mu_{\text{out}})
\]

where $W_1(\mu_{\text{in}}, \mu_{\text{out}})$ is the transportation distance defined as

\[
W_1(\mu_{\text{in}}, \mu_{\text{out}}):=\inf _{p \in \Pi\left(\mu_{\text{in}}, \mu_{\text{out}}\right)} \sum_{(e_{1}, e_{2}) \in E \times E} d_e(e_{1}, e_{2}) p(e_{1}, e_{2})
\]

Here, $d_e(e_1,e_2)$ represents the directed distance between $e_1$, an incoming edge to $u$, and $e_2$, an outgoing edge from $v$. It is defined as the minimal number of edges needed to travel from the tail of $e_1$ to the head of $e_2$. $\Pi(\mu_{\text{in}}, \mu_{\text{out}})$ represents the set of measures on $E \times E$ that project to $\mu_{\text{in}}$ and $\mu_{\text{out}}$.
\end{Def}

In \cite{Eidi2020}, it is proven that the Ollivier Ricci curvature of $(u,v)$ can be expressed as 

\[
\mathcal{O}(u,v) = \mu_0 - \mu_2 - 2\mu_3
\]

where $\mu_i$ quantifies the fraction of incoming neighbors of $u$ that are at a distance $0\leq i \leq 3$ from $v$ in an optimal transport plan. We use this expression as the definition of Ollivier Ricci curvature in the main text.

\newpage

\section{Extended results}

\subsection*{Global anti-communities seperation of the metaphor network}

\ 

\begin{longtable}{lll}
\caption{Two Anti-community category blocks and remaining categories}
\label{sitable:anticommunities}
\\
Concrete category block & Abstract category block & Remaining categories \\
\midrule
The world, the earth & Existence and its attributes & The human body \\ 
Relative position & Time & Religious groups \\ 
Occasion & Duration in time & Level land and marshes \\ 
Inhabiting and population  & Action & Stars \\ 
Buildings and inhabited places & Exemplification and specificity & Sex organs \\ 
Church government & Social communication and culture & Drinks and drinking \\ 
Land and islands & Social position & Creation \\ 
Landscape, high and low land & Moral evil & Hell \\ 
Tides, waves and flooding & Judgement & Worship \\ 
Atmosphere and weather & Enquiry and discovery & Records and monuments \\ 
Body of water & Good & Sky \\ 
Rivers and streams & Bad & Sport \\ 
Sea & Attention and inattention & The arts \\ 
Geological features & Constellations and comets & Decrease in quantity \\ 
Minerals & Answer and argument & Biological sex \\ 
Ice & Measurement of time and relative time & Completion \\ 
Life & Change and permanence & Relationship \\ 
Biological processes & Safety & Sound and video recording \\ 
Natural habitats and taxonomies & Prosperity and success & Printing and publishing \\ 
Bodily shape and strength & Disadvantage and harm & Containers \\ 
Body parts & Behaviour and conduct & Poverty \\ 
Bones, muscles and cartilage & Character and mood & Consciousness \\ 
Digestive organs & Cleverness & Workers and workplaces \\ 
Ill-health & Intelligibility & Skin \\ 
Animal categories, habitats and behaviour & Expectation and prediction & Punishment \\ 
Insects and other invertebrates & Unimportance & Sense and speech organs \\ 
Textiles & Excitement & Cessation \\ 
Clothing & Emotional suffering & Fashionableness \\ 
Sight & Anger & Death \\ 
Hearing and noise & Motivation, demotivation and persuasion & Sleep \\ 
Hardness & Wealth & Weariness \\ 
Leaking and outpouring & Taking and thieving & Dead person  \\ 
Supernatural & Command and control & Mourning and obsequies \\ 
Languages of the world & Money & Jealousy \\ 
Navigation & Social events & Peace and absence of war \\ 
Trade and commerce & Literature & Humankind \\ 
Leisure and games & Endeavour & Social attitudes \\ 
Music & Computing & Person  \\ 
Earth science & Understanding & Baby and young person \\ 
Animal bodies & Strong emotion and lack of emotion & Obtaining \\ 
Shape & Intention and planning & Relevance \\ 
Mathematics & Study of language & Uniformity and stereotypicality \\ 
Weapons and armour & Visual arts & Right and justice \\ 
Universe and space & Continuation & Degree \\ 
Size and spatial extent & Advantage & \\ 
Period of time & Order & \\ 
Wholeness & Disorder & \\ 
Deity & Greatness and intensity & \\ 
Heaven & Sufficiency and abundance & \\ 
Celestial sphere & Completeness & \\ 
Alchemy & Perception and cognition & \\ 
Chemistry & Knowledge and experience & \\ 
Direction & Language & \\ 
Number & Expression of opinion & \\ 
Military forces & Causation & \\ 
Transport & Operation and influence & \\ 
Machines & Contrast & \\ 
Lack of density & Esteem & \\ 
Day and evening & Rule and government & \\ 
Movement in a specific direction & Authority, rebellion and freedom & \\ 
Materials and fuel & Occupations and work & \\ 
Types of sport & Importance & \\ 
Heavenly body & Adversity & \\ 
Physical sensation & Vigorous action and degrees of violence & \\ 
Strength & Increase in quantity & \\ 
Wetness & Contempt & \\ 
Flowing and floating & Pleasure & \\ 
Destruction & Decision-making & \\ 
Signs and signals & Preparation and undertaking & \\ 
Travel and journeys & Inaction & \\ 
Performance arts and film & Restoration and recovery & \\ 
Planets and satellites & Similarity & \\ 
Sun & Emotion & \\ 
Astronomy & Courage & \\ 
Electromagnetism and atomic physics & Necessity and inclination & \\ 
Movement & Law & \\ 
Types of movement & Communication and disclosure & \\ 
Progressive movement & Television and broadcasting & \\ 
Birth & Relinquishing & \\ 
Birds & Age & \\ 
Canines & Ability & \\ 
Writing & Foolishness & \\ 
Cardinal points & Political office & \\ 
Hair & Education & \\ 
Groups of animals & Difficulty & \\ 
Horses and elephants & Psychology & \\ 
Food and eating & Philosophy & \\ 
Touch & Solitude and social isolation & \\ 
Dirtiness & Failure & \\ 
Weakness & Sequence & \\ 
Softness & Insufficiency & \\ 
Bad condition & Intellect & \\ 
Rate of movement and swift movement & Stupidity & \\ 
Impulse & Intellectual weakness & \\ 
Impact & Belief and opinion & \\ 
Part-whole relationships & Willingness and desire & \\ 
Industry & Information and advertising & \\ 
Tools and equipment for work & Air and space travel & \\ 
Absence of life & Incompleteness & \\ 
Biology & Mind & \\ 
Wild and fertile land & Moderateness and smallness of quantity & \\ 
Lakes and pools & Lack of understanding & \\ 
Angel & Pride & \\ 
War and armed hostility & Humility & \\ 
Region of the earth & Licentiousness & \\ 
Pigs & Drug use & \\ 
Weight, heat and cold & Easiness & \\ 
Physics and mechanics & Foolish person & \\ 
Place and position & Disbelief and uncertainty & \\ 
Immobility and restlessness & Giving & \\ 
Correlation & Speaking & \\ 
Equivalence & Morality and immorality & \\ 
Possessing and lacking & Correspondence and telecommunications & \\ 
Railways & Sexual relations & \\ 
Hunting and fishing & Truth and falsity & \\ 
Solidity and density & Pity and pitilessness & \\ 
Religious places and artefacts & Composure & \\ 
Representation & Will and personal choice & \\ 
Races and nations & Insignia and heraldry & \\ 
Unctuousness & Memory, commemoration and revocation & \\ 
Respiratory system & Hatred and hostility & \\ 
Healing and treatment & Curse & \\ 
Classes of mammals & Social discord and harmony & \\ 
Other clawed mammals & Faith & \\ 
Awake & Journalism & \\ 
Dryness & Reason and argument & \\ 
Distance & Politics & \\ 
Transference & Virtue & \\ 
Measurement of length & Air and sea hostilities & \\ 
Measurement of volume & Mental health & \\ 
Measuring instrument & Beauty and ugliness & \\ 
Providing and storing & Female person & \\ 
Farming & Imagination & \\ 
Bodily tissue & Secrecy and concealment & \\ 
Space & Love and friendship & \\ 
Internal organs & Wisdom & \\ 
Bodily secretion & Marriage & \\ 
Cleanness & Slow action and degrees of caution & \\ 
Bodily excretion & Society & \\ 
Smell & Thought & \\ 
Vascular system & Lack of knowledge & \\ 
Killing & Fear & \\ 
Taste & Chance & \\ 
Liquid & Suitability of time & \\ 
Guilt & Tastelessness & \\ 
The brain and nervous system & Retaining & \\ 
Human anatomy & Cacti, ferns, moss and algae & \\ 
Plants & Invigoration & \\ 
Granular texture & Aesthetics and good taste & \\ 
Air & Refusal & \\ 
Artificial light & Frequency & \\ 
Darkness & Occurrence & \\ 
Colour  & Difference & \\ 
Night & & \\ 
Loss & & \\ 
Water & & \\ 
Manner of death & & \\ 
Cause of death & & \\ 
Burial and cremation & & \\ 
Health & & \\ 
Flickering and glowing light & & \\ 
Individual colours & & \\ 
Good condition & & \\ 
Measurement of weight & & \\ 
Gas & & \\ 
Imitation & & \\ 
Devil & & \\ 
Wrong and injustice & & \\ 
Animals & & \\ 
Male person & & \\ 
Old person  & & \\ 
Amphibians & & \\ 
Reflection & & \\ 
Pattern and variegation & & \\ 
Adult and middle-aged person & & \\ 
Named regions of earth & & \\ 
Dance & & \\ 
Fish & & \\ 
Reptiles & & \\ 
Felines & & \\ 
Ruminants & & \\ 
Light & & \\ 
Family members and genealogy & & \\ 
Fireworks & & \\ 
Slow movement & & \\ 
Astrology & & \\ 
Rhinoceroses and general ungulates & & \\ 
Bats, aardvarks, flying lemurs and tree-shrews & & \\ 
Primates & & \\ 
Zoology and taxidermy & & \\ 
Variety & & \\ 
Kinship and relationship & & \\ 
Kinship group and family & & \\ 
Books & & \\ 
Flowers and grasses & & \\ 
Trees and shrubs & & \\ 
Semi-fluidity & & \\ 
Cultivated plants & & \\ 
Weeds & & \\ 
Botany & & \\ 
Repetition & & \\ 
Poison & & \\ 
Fine and coarse texture & & \\ 
Humidity & & \\ 
Execution and performance & & \\ 
Illumination & & \\ 
Natural light & & \\ 
Transparency and opacity & & \\ 
Sameness & & \\ 
Naming & & \\ 
Accompaniment & & \\ 
Measurement & & \\ 
Measurement of area & & \\ 
Quantity & & \\ 
Indifference & & \\ 
Gratitude and ingratitude & & \\ 
Reading & & \\
\bottomrule
\end{longtable}

\subsection*{Hierarchical Cluster Analysis (HCA)}
The HCA algorithm produced only four dendrograms (binary trees). This remarkably small number of dendrograms, resulting from ties in proximity \cite{MacCuish2001,Leal2016}, shows the stability of the global semantic role structure of categories. 

\begin{figure}
    \centering
    \includegraphics[width=\linewidth]{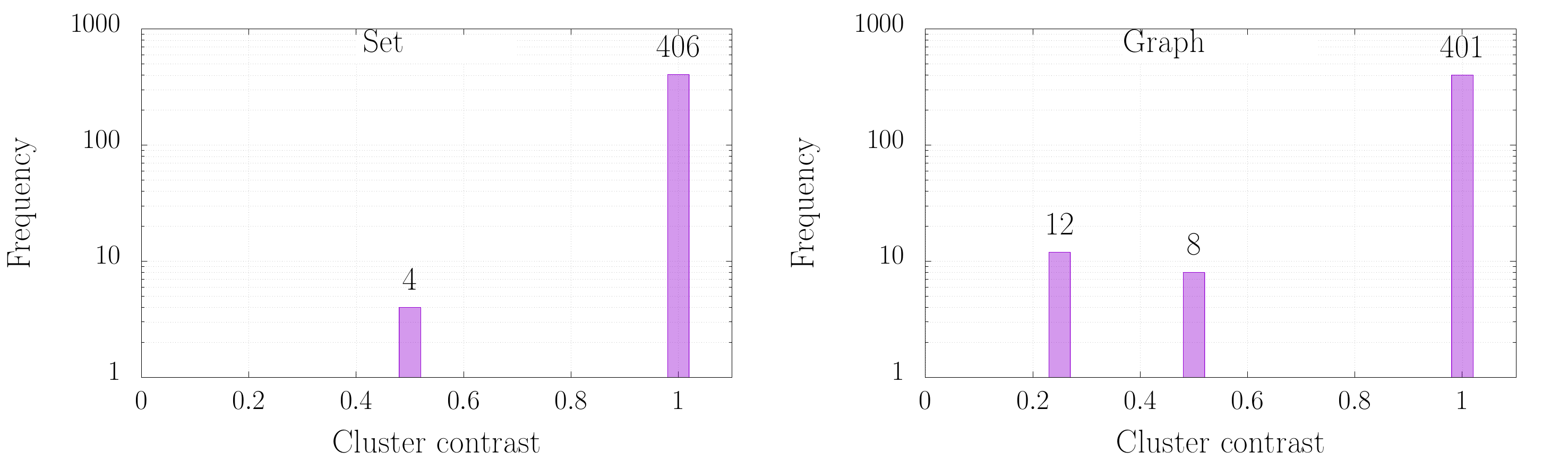}
    \caption{Distribution of cluster contrast values for clusters within the four dendrograms obtained from hierarchical cluster analysis of categories in the metaphor network, using $d(c,c') = \frac{\#(N_{\text{in}}(c) \ominus N_{\text{in}}(c'))}{\#(N_{\text{in}}(c) \cup N_{\text{in}}(c'))} + \frac{\#(N_{\text{out}}(c) \ominus N_{\text{out}}(c'))}{\#(N_{\text{out}}(c) \cup N_{\text{out}}(c'))}$ as a distance between categories $c$ and $c'$, and Ward's grouping methodology.}
    \label{fig:dendron1}
\end{figure}

\begin{figure}
    \centering
    \includegraphics[width=0.49\linewidth]{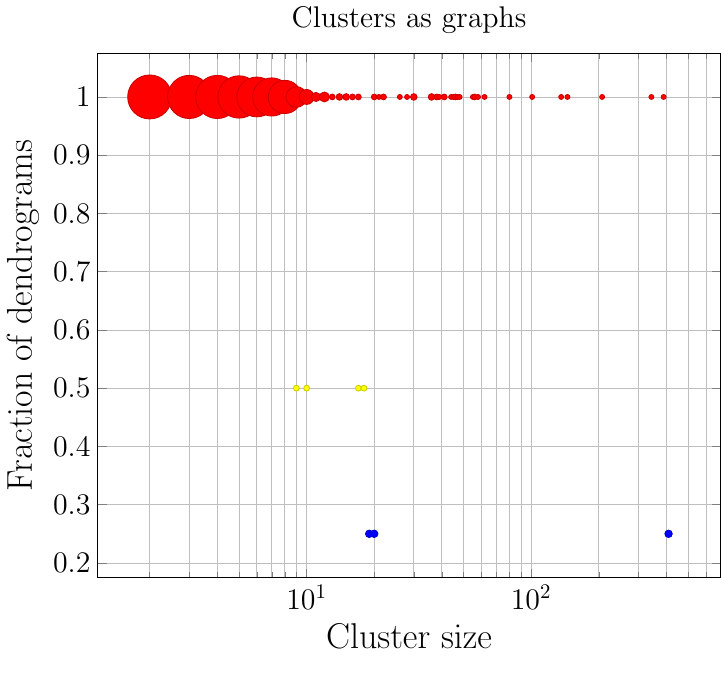}
    ~
    \includegraphics[width=0.49\linewidth]{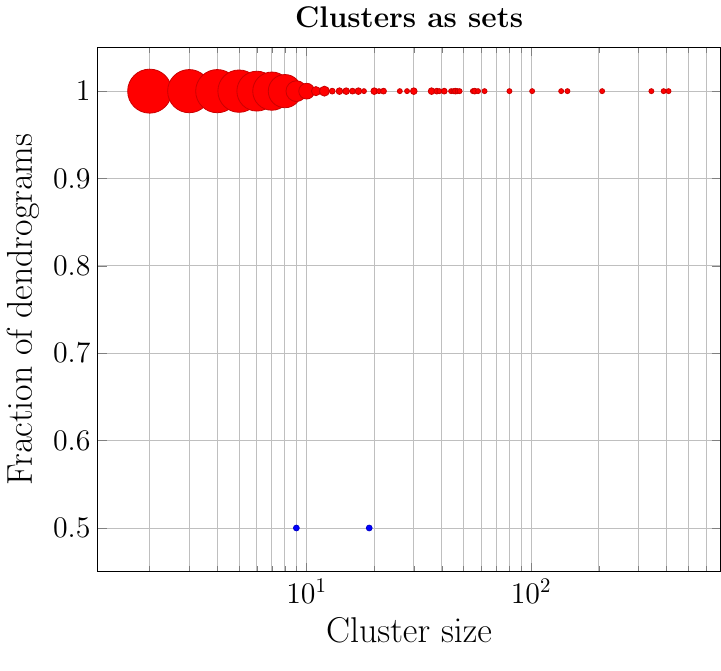}
    \caption{The Hierarchical Cluster Analysis (HCA) algorithm ran on the collection of categories in the metaphor network generated four different dendrograms.  Most cluster consistently appeared in all four dendrograms.  These scatter plots show the distribution of pairs (cluster size, fraction of dendrograms) corresponding to the clusters found within the four dendrograms.  The size of a cluster is defined as its number of leaves, that is, the number of categories within the cluster.  The ball size in the plot represents the frequency of occurrence for a specific pair  (cluster size, fraction of dendrograms).  The scatter plot on the left defines clusters as subtrees of the dendrograms, while the scatter plot on the right defines clusters as the collection of leaves of subtrees of the dendrograms.  The HCA algorithm was ran using: $d(c,c') = \frac{\#(N_{\text{in}}(c) \ominus N_{\text{in}}(c'))}{\#(N_{\text{in}}(c) \cup N_{\text{in}}(c'))} + \frac{\#(N_{\text{out}}(c) \ominus N_{\text{out}}(c'))}{\#(N_{\text{out}}(c) \cup N_{\text{out}}(c'))}$ as a distance between categories $c$ and $c'$, and Ward's grouping methodology.}
    \label{fig:dendron2}
\end{figure}

The four dendrograms encompass a total of 421 distinct graph clusters, defined as all the different subtrees that can be extracted from the four dendrograms. In Figure \ref{fig:dendron1}, it can be observed that 401 graph clusters, the majority of them (95.3\%), consistently appear as subtrees in all four dendrograms. The remaining twenty graph clusters, which are not present as subtrees in all four dendrograms, can be categorized as follows (Figure \ref{fig:dendron2}): the four entire dendrograms (cluster size equal to 415), eight relatively small-sized subtrees (approximately 20 leaves) appearing in two out of the four dendrograms, and twelve subtrees ranging from 8 to 18 leaves, which appear in only one dendrogram.

Additionally, considering clusters as sets of leaves rather than graph structures provides an alternative perspective \cite{Leal2016}. For instance, although the binary trees $((a,b),c)$ and $(a, (b,c))$ are distinct graph structures, they share the same set of leaves, namely, ${a,b,c}$. Similarly, the four dendrograms obtained share the same set of leaves, making them stable clusters as sets but not as graphs. In this context, we computed the clusters based on their sets of leaves, disregarding the hierarchical graph structure, and determined that there are 410 set clusters, which is 11 fewer than the number of graph clusters. Furthermore, 406 set clusters (99.0\%) consistently appear as subclusters in all four dendrograms, while the remaining four set clusters appear in two dendrograms each.

In summary, the results obtained from both graph and set clusters indicate that despite being merged differently during the grouping process, leading to distinct graph structures, the semantic roles exhibit remarkable stability.

\end{document}